# ALL-PET: A Low-resource and Low-shot PET Foundation Model in Projection Domain

Bin Huang[2,#], Kang Chen[2,#], Bingxuan Li[3], Huafeng Liu[4], Qiegen Liu [1,*]

[1]School of Information Engineering, Nanchang University, Nanchang 330031, China

[2] School of Mathematics and Computer Science, Nanchang University, Nanchang 330031, China

[3] Institute of Artificial Intelligence, Hefei Comprehensive National Science Center, Hefei 230001, China

[4] College of Optical Science and Engineering, Zhejiang University, Zhejiang 310058, China

[#]The authors contributed equally to this work

[*] Corresponding author: liuqiegen@ncu.edu.cn

**Building large-scale foundation model for PET imaging is hindered by limited access to labeled data and insufficient computational resources. To overcome data scarcity and efficiency limitations, we propose ALL-PET, a low-resource, low-shot PET foundation model operating directly in projection domain. ALL-PET leverages a latent diffusion model (LDM) with three key innovations. First, we design a Radon mask augmentation strategy (RMAS) that generates over 200,000 structurally diverse training samples by projecting randomized image-domain masks into sinogram space, significantly improving generalization with minimal data. This is extended by a dynamic multi-mask (DMM) mechanism that varies mask quantity and distribution, enhancing data diversity without added model complexity. Second, we implement positive/negative mask constraints to embed strict geometric consistency, reducing parameter burden while preserving generation quality. Third, we introduce transparent medical attention (TMA), a parameter-free, geometry-driven mechanism that enhances lesion-related regions in raw projection data. Lesion-focused attention maps are derived from coarse segmentation, covering both hypermetabolic and hypometabolic areas, and projected into sinogram space for physically consistent guidance. The system supports clinician-defined ROI adjustments, ensuring flexible, interpretable, and task-adaptive emphasis aligned with PET acquisition physics. Experimental results show that ALL-PET achieves high-quality sinogram generation using only 500 samples, with performance comparable to models trained on larger datasets. ALL-PET generalizes across tasks including low-dose reconstruction, attenuation correction, delayed-frame prediction, and tracer separation, operating efficiently with memory use under 24GB.**

## Introduction

Foundation models have made notable progress in medical imaging by leveraging large-scale neural networks pretrained on diverse datasets[1-3]. They have achieved notable results across various clinical applications, including MRI[4–5], CT[6–8], ophthalmology[9], radiology[10], and pathology[11–14]. This performance is largely due to their ability to learn generalizable features from large and heterogeneous data. Foundation models typically require a high number of parameters to support broad generalization, which increases computational cost and reduces interpretability[15]. Most existing medical foundation models [16-19] operate in the image domain. For PET imaging, this is suboptimal, as it discards projection data that contain important physical priors essential for accurate tracer quantification. A projection domain approach may better preserve these properties and offer improved transparency for clinical use.

Medical foundation models still require large and diverse datasets for pretraining[20-25]. However, in clinical settings, access to such data is limited due to privacy concerns, ethical constraints, and the high cost of expert annotations[26-29]. Medical imaging data also show high variability across patients and modalities, making it difficult to construct standardized datasets[30-31]. These challenges are particularly significant in PET imaging, where raw projection data are complex and are not widely available. Therefore, there is an urgent need for low-shot foundation models that can perform well with limited data and still retain modality specific physical priors.

In addition to requiring large and diverse datasets, most foundation models also rely on hundreds of millions of parameters[38-39], resulting in computationally demanding architectures. This makes them difficult to deploy in multi-center clinical settings or low-resource environments[40-42]. PET imaging, in particular, would benefit from low-resource foundation models[43-46] that require fewer resources while still preserving task-relevant physical priors. There is a growing need for models that can achieve generalizable performance under both data-limited and computation-constrained scenarios.

A key limitation of current medical foundation models is their confinement to the image domain. Although this is compatible with most clinical workflows, it excludes projection domain, which represents the raw measurement space in PET imaging. Projection domain, typically represented in Radon space, retains essential physical priors, including attenuation effects, anatomical geometry, and view-angle dependencies[47-48]. These features are often degraded or lost during image reconstruction[49]. Ignoring projection domain can lead to outputs that appear anatomically realistic but lack physical consistency, reducing both diagnostic reliability and model interpretability. Moreover, the absence of explicit physical constraints in deep architectures limits transparency, which is an essential requirement for clinical adoption in nuclear medicine.

In light of these limitations, there is a clear need for a PET foundation model that is low-resource, low-shot, and grounded in projection domain. Such a model should be capable of capturing physical priors directly from sinograms while maintaining interpretability and adaptability across clinical tasks. To our knowledge, no prior foundation model fulfills these criteria in the PET field. Thus, we introduce ALL-PET, the first low-resource and low-shot foundation model tailored for PET projection data. ALL-PET directly operates in the sinogram domain using latent diffusion modeling techniques, eliminating the need for large image domain pretraining. Our research demonstrates ALL-PET's ability to generate high-quality synthetic sinograms and various downstream tasks such as low-dose PET reconstruction, self-attenuation correction, delayed-frame prediction, tracer conversion and tracer separation. We also show that ALL-PET supports integration with computed tomography (CT) data for brain and trunk synthesis, highlighting its potential for continuous knowledge acquisition and cross-modality learning. Compared to recent foundation models[50-52], ALL-PET achieves competitive results while maintaining low-resource and low-resource, marking a significant step toward clinically deployable PET foundation models.

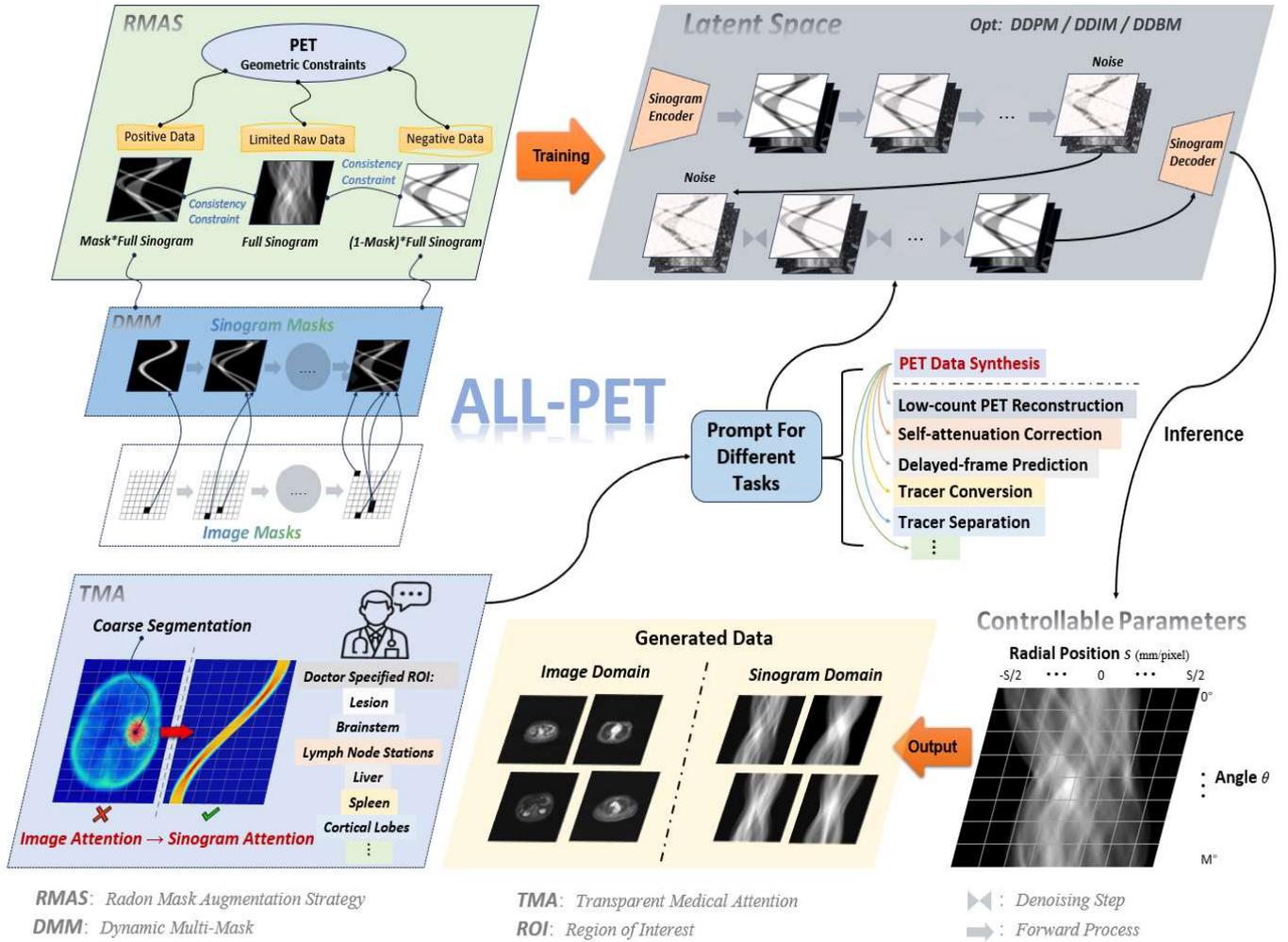

Fig.1. Overview of the ALL-PET foundation model and training/inference pipeline. RMAS and DMM are used for PET sinogram synthesis while TMA is used for downstream tasks. By incorporating DDPM/DDIM/DDBM into the latent diffusion, ALL-PET can be utilized as supervised and unsupervised model. Prompt is used for generating different tracer or anatomical regions and controllable parameters can output specified size of the sinograms.

## Results
### Overview of ALL-PET

To achieve low-resource and low-short features in projection domain of PET, we designed the ALL-PET foundation model as outlined below. The overall training and inference process of the ALL-PET is illustrated in Fig. 1. The core framework is built upon latent diffusion model (LDM) and enhanced with three key innovations. Firstly, a Radon mask augmentation strategy (RMAS) is introduced to exponentially expand training samples from limited datasets. By applying 64×64 image-domain mask blocks and projecting them into sinogram space via Radon transform, ALL-PET generates over 200,000 structurally diverse Radon masks. Each mask encodes spatial position and angular geometry, significantly enhancing the model's generalization and adaptation to unseen data with minimal real samples. Furthermore, this strategy is extended by implementing a dynamic multi-mask mechanism (DMM), where the number, size, and spatial distribution of masks vary dynamically within each training sample. This exponentially increases data diversity without adding model complexity, effectively enhancing ALL-PET's low-resource characteristics and robustness under limited data conditions.

Secondly, a physically-constrained modeling framework is proposed based on positive/negative mask regularization. The model enforces strict geometric consistency through mathematically defined relationships: $ch_1 = mask \times ch_2$ and $ch_3 = (1 - mask) \times ch_2$, where $ch_2$ denotes the full sinogram. These constraints act as zero-cost implicit regularizers, embedding prior knowledge of sinogram structure without increasing model parameters. Compared to traditional low-resource methods that reduce network capacity, ALL-PET leverages intelligent data design and physics-guided learning to maintain generation quality with significantly reduced model complexity. DMM creates massive, randomized geometric constraints for ALL-PET. The network is forced to continuously learn and adapt to new data relationships defined by different mask configurations, all within a unified LDM. By leveraging the complexity of intelligent data constraints, ALL-PET replaces the need for excessive model parameter complexity, achieving high-quality results with a low-resource design.

Thirdly, we further introduce transparent medical attention (TMA), a parameter-free and geometry-driven attention mechanism designed for PET projection data. This method constructs interpretable attention maps based on PET imaging physics and sinogram geometry, enhancing the contribution of relevant ROI in raw projection data without introducing additional trainable parameters. In this study, attention maps are primarily used for different downstream tasks. General lesion regions are first identified via thresholding of fast, low-resolution FBP reconstructions, covering both hypermetabolic and hypometabolic abnormalities. These lesion-focused attention masks are forward projected into sinogram space, creating explicit, physically consistent guidance for various downstream tasks. Importantly, the system is designed with clinical flexibility, physicians can manually adjust or redefine ROI regions based on specific diagnostic needs, seamlessly integrating expert knowledge into the attention mechanism without disrupting model integrity. This strategy ensures that ALL-PET maintains low-resource efficiency while providing interpretable, task-adaptive, and clinically customizable ROI emphasis aligned with PET acquisition physics.

**Data collection and experimental design**

The experiment data is generated by the DigitMI 930 PET/CT scanner. This scanner is developed by RAYSOLUTION Healthcare Co., Ltd, and incorporate state-of-the-art all-digital PET detectors. The PET scanner has an axial field-of-view (AFOV) of 30.6 cm within an 81 cm ring diameter.

Patient data: A comprehensive data set is utilized, consisting of 14 patients. Each patient undergoes a scan ranging from 4 to 8 beds, with a complete sampling scan time of 45 seconds to 3 minutes per bed. From these patients, 500 sinograms are randomly sampled for training and 100 sinograms for testing, which are utilized for each synthetic tasks and downstream tasks.

For synthetic tasks, the training dataset consists of three radiotracers, $^{18}$F-FDG, $^{18}$F-DOPA, and $^{68}$Ga-PSMA, as well as different anatomical regions, including the brain and trunk (with the trunk further divided into chest and abdomen). For the foundation model training, 500 sinograms per tracer or per anatomical region are used to ensure balanced learning across different conditions.

For downstream tasks, the dataset includes paired low-dose and normal-dose data, pre- and post-attenuation correction, first-scan and delayed-scan synthesis, dual-tracer separation, and tracer-to-tracer conversion. For each downstream task, 500 paired sinograms are used for training and 100 paired sinograms for testing to enable comprehensive evaluation and maintain consistency across different task settings. This study is approved by the institutional review board of the Beijing Friendship Hospital, Capital Medical University, Beijing, China. The approval number is 2022-P2-314-01.

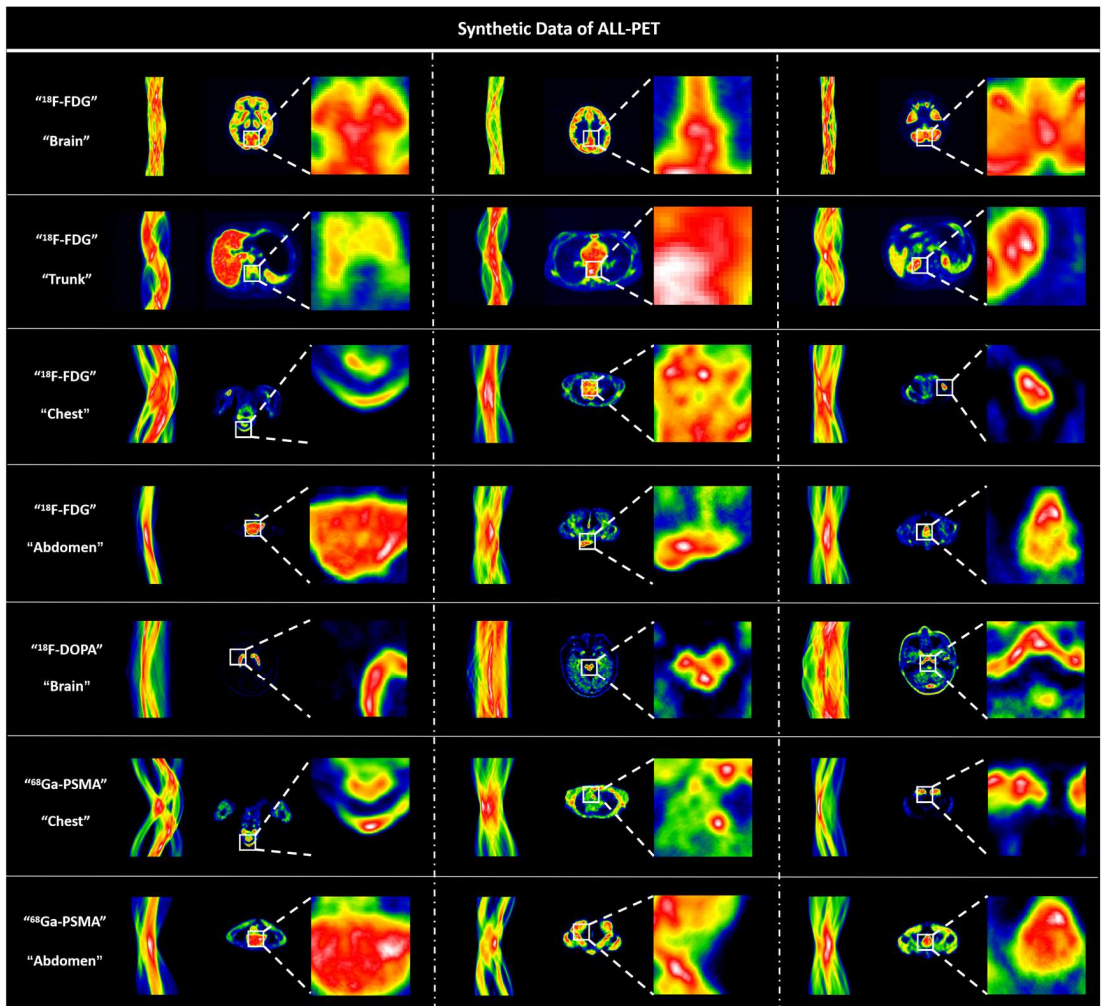

Fig. 2. Synthetic projection domain data generated by ALL-PET and their corresponding reconstructed PET images, shown for multiple radiotracers e.g., $^{18}$F-FDG, $^{18}$F-DOPA, $^{68}$Ga-PSMA and different anatomical regions e.g., Brain, Trunk, Chest, Abdomen.

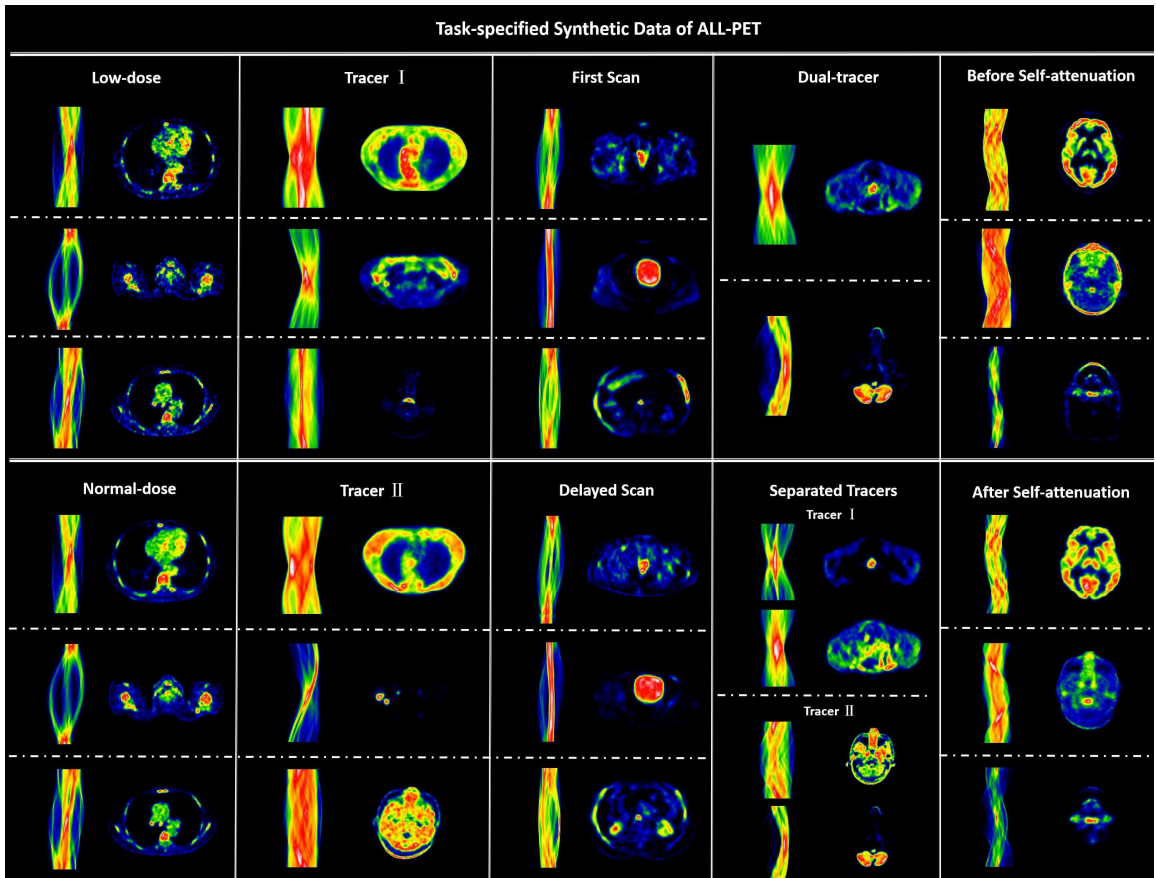

Fig. 3. Task-specified paired synthetic data generated by ALL-PET, showing projection domain data and their corresponding reconstructed PET images for five representative downstream tasks e.g., low-dose and normal-dose data, before and after attenuation correction, first scan and delayed scan synthesis, dual-tracer separation and tracer conversion.

In our experiment, ALL-PET is trained by the AdamW algorithm with a learning rate of $2.0 \times 10^{-6}$. Synthetic data are generated using denoising diffusion implicit model (DDIM) sampling, where the shape of the sampled data is 2, 3, 512, 512, the eta value is set to 1.0, and the number of sampling steps is 250. This method is implemented in Python and PyTorch on a personal workstation equipped with a GPU card NVIDIA 3090 24GB. In the generation stage, ALL-PET conducts 72 steps of DDIM sampling. After the sampling is completed, the training weights are restored to complete the process of generating data in projection domain.

**Synthesis data evaluation**

To evaluate the generative performance of ALL-PET, we synthesized sinogram data from randomly sampled text prompts covering various tracer–anatomy combinations, such as "brain with $^{18}$F-FDG," "trunk with $^{18}$F-DOPA," and "chest or abdomen with $^{68}$Ga-PSMA." As illustrated in Fig. 2, the generated sinograms reflect realistic anatomical structures and tracer uptake distributions, closely aligned with clinical expectations. Both global anatomical consistency and local uptake fidelity are preserved across different anatomical regions.

We further examined task-specific synthesis scenarios. Fig. 3 presents paired synthetic data for five representative downstream tasks, low-dose and normal-dose data, before and after attenuation correction, first scan and delayed scan synthesis, dual-tracer separation and tracer conversion. In all cases, anatomical and uptake fidelity are well preserved across both domains, underscoring the consistency of ALL-PET.

A qualitative method comparison is displayed in Fig. 4(a), where VAE[51], GAN[52], SGM[53], and ALL-PET outputs are juxtaposed in both domains. Under the 500-sample regime, the VAE recovers only coarse contours with most fine detail missing; GANs reconstruct some structural detail but exhibit unstable or unrealistic textures; SGM produces markedly better fine structure than GAN, yet still falls short of ALL-PET. ALL-PET consistently preserves fine anatomical features and tracer-specific uptake patterns that are lost or corrupted by other methods.

Quantitative evaluations appear in Fig. 4(b–c). Projection and image domain similarity are assessed using Fréchet Inception Distance (FID), Kernel Inception Distance (KID), and Inception Score (IS). ALL-PET attains the lowest FID and KID values across both domains, indicating superior distributional alignment with real sinograms and reconstructed images. In the reconstructed image domain, ALL-PET achieves FID values of 68.01 for 18F-FDG brain, 65.21 for $^{18}$F-FDG trunk, 131.65 for $^{18}$F-DOPA, and 304.04 for $^{68}$Ga-PSMA, with corresponding KID values of 0.043, 0.033, 0.087, and 0.084, respectively. These results indicate that ALL-PET delivers higher visual fidelity, greater sample diversity, and closer statistical agreement with real data than the baseline generative models under low-shot conditions.

Overall, both qualitative and quantitative evidence demonstrate that ALL-PET reliably synthesizes clinically meaningful and anatomically diverse projection domain data, and that this fidelity is preserved after image reconstruction, supporting its utility for downstream PET tasks in data-scarce settings.

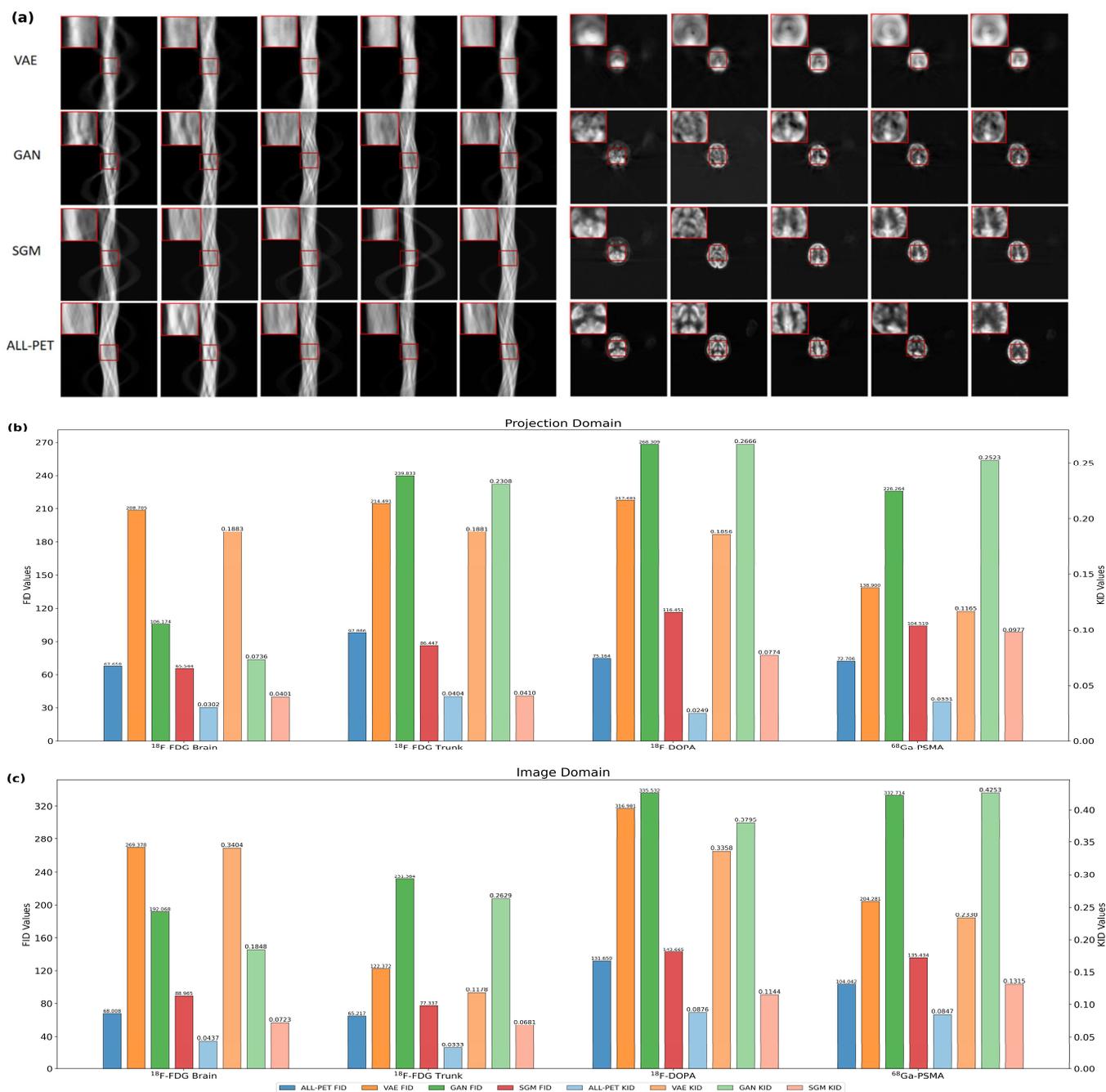

Fig.4. Qualitative and quantitative comparison of VAE, GAN, SGM, and ALL-PET in projection and image domains. (a) visual juxtaposition of generated outputs from each method, with red boxes denoting regions enlarged for detailed inspection; (b) projection domain FID and KID values for three radiotracers: [18]F-FDG, [18]F-DOPA and [68]Ga-PSMA; (c) image domain FID and KID values for the same radiotracers.

**Downstream task evaluation**

Unlike conventional methods in which downstream PET tasks are predominantly executed in the image domain, ALL-PET performs all task-specific operations directly in projection domain, thereby minimizing error propagation that typically arises during intermediate image reconstruction. This dual-domain design ensures that the intrinsic statistical properties of projection data are preserved while still enabling clinically meaningful image-level outputs. By integrating multiple diffusion paradigms, the framework is highly flexible: supervised training can be realized through the denoising diffusion bridge model (DDBM), while unsupervised strategies can be employed using denoising diffusion probabilistic model (DDPM) or DDIM, depending on the task and data availability.

Fig. 5 illustrates representative outcomes for four clinically relevant downstream applications, with projection domain results shown in panel (a) and the corresponding image domain reconstructions in panel (b). For each task, training is conducted with only 500 paired sinograms using the DDBM setting, thereby evaluating the capacity of ALL-PET under data-scarce conditions.

For low-dose PET reconstruction, input sinograms correspond to merely 1/100 of the standard acquisition counts. Despite the extreme photon sparsity, ALL-PET generates reconstructions with markedly enhanced contrast, reduced stochastic noise, and sharper delineation of anatomical structures. Quantitative evaluations confirm these improvements, with significantly higher peak signal-to-noise ratio (PSNR), better structural similarity index (SSIM), and lower root mean squared error (RMSE), indicating both visual and numerical superiority.

For attenuation artifact correction, projection data intentionally corrupted by simulated attenuation artifacts are generated by ALL-PET into corrected forms, effectively removing projection-level distortions. The corresponding reconstructed images exhibit anatomically consistent uptake distributions and structural alignment, free of attenuation-induced degradations that typically compromise diagnostic utility.

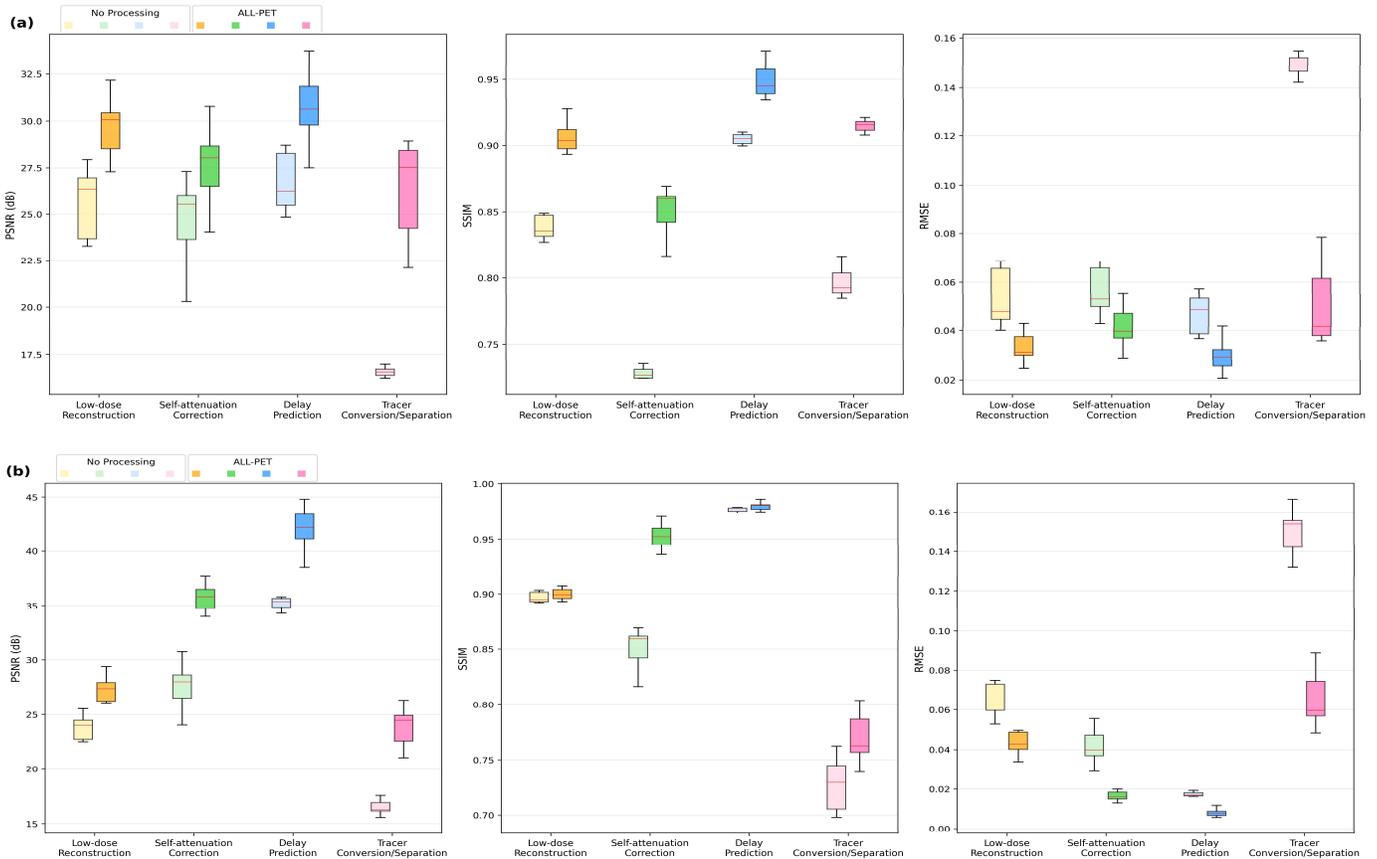

Fig.5. Boxplots of different downstream tasks e.g., low-dose reconstruction (1/100), attenuation correction, delayed-frame synthesis, tracer separation and conversion (a) demonstrating projection domain generation while (b) corresponding to the reconstructed PET images with PSNR, SSIM, MSE.

In dynamic PET scenarios, ALL-PET is evaluated for predicting delayed-frame acquisitions. The framework successfully generates temporally consistent tracer distributions, producing synthetic frames that closely match ground truth dynamics across multiple time points. This demonstrates the model's ability to capture physiologically relevant uptake patterns and tracer kinetics, which are critical for functional imaging and time-resolved analysis.

Finally, ALL-PET is tested on dual-tracer scenarios where overlapping signals present a significant challenge for conventional methods. The model generates well-separated tracer-specific distributions with high quantitative fidelity, ensuring accurate disentanglement of mixed signals. In addition, tracer-to-tracer conversion is achieved while preserving both anatomical consistency and realistic uptake patterns. These results highlight the generalizability of ALL-PET, enabling it not only to separate but also to translate tracer distributions in clinically meaningful ways.

Taken together, these results demonstrate the versatility and robustness of ALL-PET across a diverse set of PET-specific downstream tasks. By performing operations directly in projection domain while maintaining image domain fidelity, the framework enables accurate reconstruction, artifact correction, temporal synthesis, and tracer disentanglement, all under limited training data conditions, thereby advancing the practical utility of generative diffusion models in PET imaging.

**Downstream application using synthetic data**

We further validated the effectiveness of ALL-PET–generated synthetic data in improving downstream model training for attenuation correction. As illustrated in Fig. 5, the incorporation of synthetic data (SD) from ALL-PET led to consistently superior quantitative performance compared with baseline methods, including U-Net[57] and CycleGAN[56]. Specifically, ALL-PET with synthetic data achieved higher PSNR, improved SSIM, and lower RMSE values across evaluation sets. While U-Net, as a supervised framework, generally outperformed CycleGAN, its improvements were still limited by reliance on paired training data. CycleGAN, although designed for unpaired learning, produced over-smoothed results with visible loss of structural detail, further emphasizing the gap with ALL-PET.

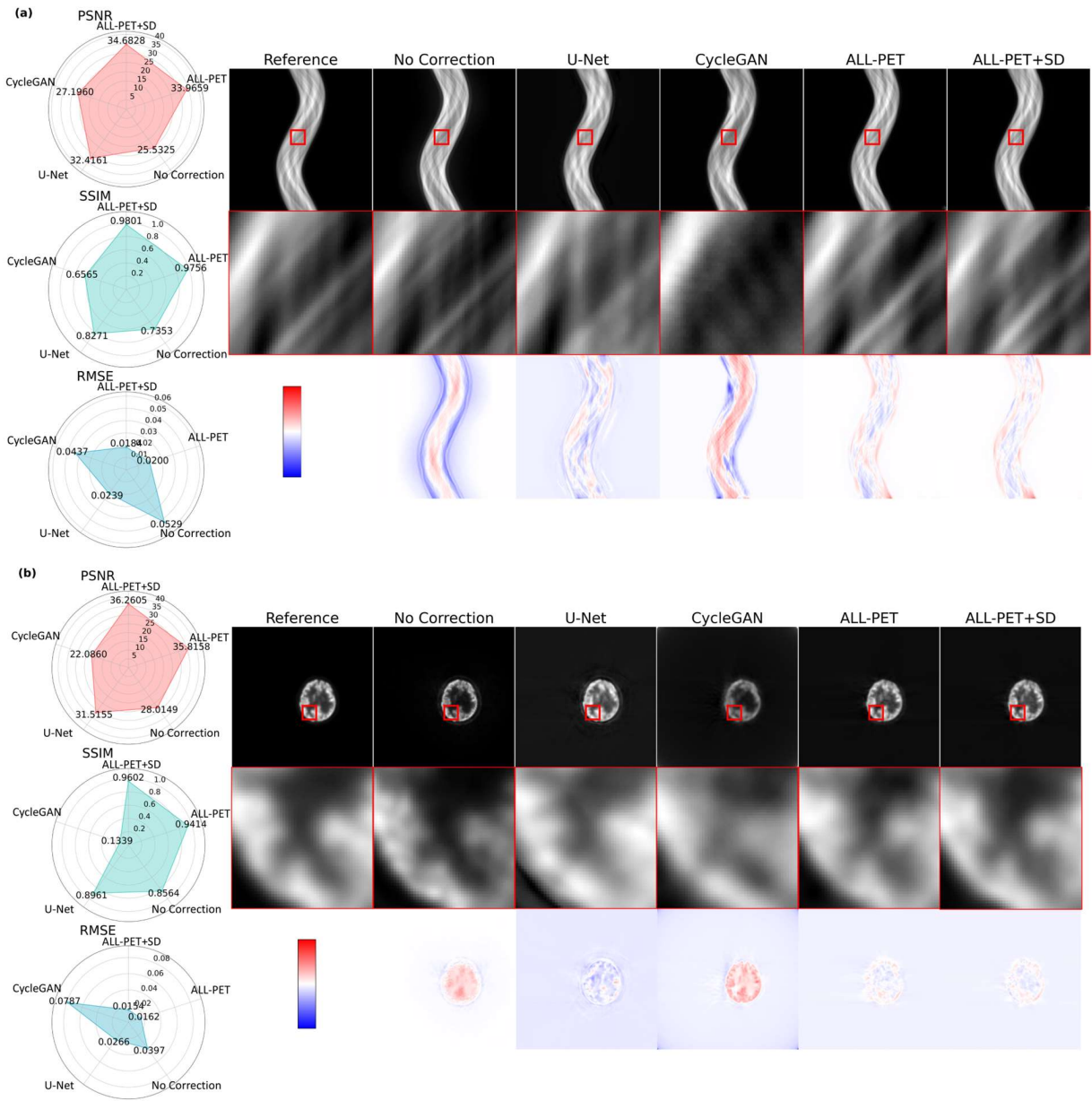

Fig. 6. Qualitative and quantitative comparison of No Correction, U-Net, CycleGAN, ALL-PET, and ALL-PET+SD in projection and image domains. Visual juxtaposition of generated outputs from each method, with red boxes denoting regions enlarged for detailed inspection are shown on the right. (a) represents projection domain; (b) represents image domain with PSNR, SSIM, RMSE on the left.

Importantly, despite being trained with only 500 sinogram pairs, ALL-PET demonstrated competitive results in low-resource settings, underscoring its efficiency in data-scarce environments. Beyond direct reconstruction quality, the generation of additional attenuation-corrected samples by ALL-PET and their integration into the training pipeline provided clear benefits. Radar plot analyses confirmed that augmenting the training dataset with synthetic attenuation-corrected data yielded measurable gains in PSNR, SSIM, and RMSE, indicating improved stability and robustness of the trained models.

These findings highlight that synthetic data generated by ALL-PET serve not only as an effective augmentation strategy but also as a means of enhancing the generalizability of existing networks. By bridging the gap between supervised and unpaired paradigms, ALL-PET strengthens model robustness in attenuation correction and facilitates reliable performance even under low-resource or data-constrained scenarios.

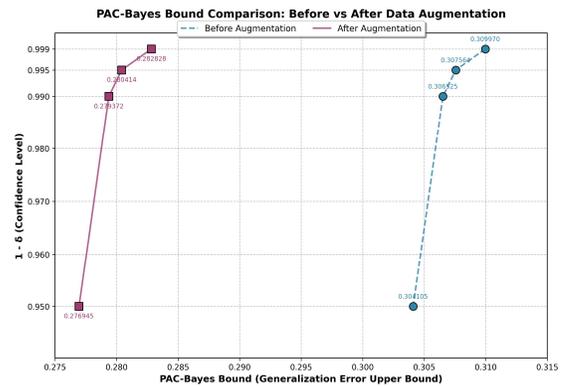

Fig. 7. PAC-Bayes upper bound comparison before and after ALL-PET data augmentation algorithm.

**PAC-Bayes bound analysis**

In the ALL-PET setting, a parametric generator is treated as a hypothesis indexed by a parameter vector. Concretely, the generator is specified by a parameter vector $\theta$:

$$\theta = (\theta_1, \theta_2, \ldots, \theta_d) \in \mathbb{R}^d, \quad (1)$$

where $\theta$ is a $d$-dimensional real-valued vector whose components correspond to the model parameters such as the weights and biases of the generative neural network. In the unconditional generation regime considered here there is no input $x$, synthetic PET sinograms are produced by sampling latent noise and applying the generator controlled by $\theta$. The observed dataset is a finite collection of independently observed PET sinograms:

$$S = \{y_1, y_2, \ldots, y_n\}, \quad (2)$$

where each $y_i$ is a sinogram drawn independently from the unknown data distribution $\mathcal{D}$ over sinograms.

To quantify the quality of a generator parameter $\theta$ on an observed sinogram $y$, a per-sample loss function $\ell(\theta; y)$ is employed to measure the discrepancy between generated and observed data. In order to apply PAC-Bayes concentration results, the per-sample loss is required to be bounded in the unit interval; hence, the natural loss like mean squared error is rescaled and clipped as $0 \le \ell(\theta; y) \le 1$ for every $\theta$ and $y$.

The true expected risk of the parameter vector $\theta$ is defined as the expectation of the per-sample loss under the data distribution $D$:

$$R(\theta) = \mathbb{E}_{y \sim D}[\ell(\theta; y)] \quad (3)$$

Given the finite sample $S$ of size $n$, the empirical risk of $\theta$ is the sample average:

$$\hat{R}(\theta) = \sum_{i=1}^{n} \ell(\theta; y_i) / n \quad (4)$$

Randomized (Gibbs) predictors are introduced by placing probability measures on the parameter space. Let $P$ denote a fixed prior probability measure on the parameter space, chosen independently of the sample $S$. Let $Q$ denote any posterior probability measure on the parameter space, possibly depending on training data and checkpoints. The Gibbs empirical and true risk under $Q$ are the corresponding expectations:

$$\hat{R}(Q) = \mathbb{E}_{\theta \sim Q}[\hat{R}(\theta)], \quad R(Q) = \mathbb{E}_{\theta \sim Q}[R(\theta)] \quad (5)$$

The Kullback–Leibler divergence between the posterior $Q$ and the prior $P$ is written explicitly as:

$$KL(Q \| P) = \mathbb{E}_{\theta \sim Q}[\ln(Q(\theta) / P(\theta))], \quad (6)$$

where densities or mass functions are taken with respect to the same dominating measure. Because the concentration step is most naturally expressed using the binary (Bernoulli) relative entropy, the binary Kullback–Leibler divergence between two probabilities $p, q \in [0,1]$ is defined by:

$$kl(p \| q) = p\ln(p/q) + (1-p)\ln[(1-p)/(1-q)] \quad (7)$$

The derivation begins with a concentration bound for a fixed parameter vector $\theta$. For this fixed $\theta$, define the independent random variables $Z_i = \ell(\theta; y_i)$, $i = 1, \ldots, n$, so that $0 \le Z_i \le 1$ and $\mathbb{E}[Z_i] = R(\theta)$. The empirical risk is $\hat{R}(\theta) = \sum_{i=1}^{n} Z_i / n$. A standard large-deviations (Chernoff/Cramér) argument for bounded variables yields the fundamental moment bound:

$$\mathbb{E}_S[\exp(n \cdot kl(\hat{R}(\theta) \| R(\theta)))] \le n + 1, \quad (8)$$

where $\mathbb{E}_S$ denotes expectation over the random draw of the sample $S$. Markov's inequality applied to this moment bound implies that for any $\varepsilon > 0$:

$$P_S r \{kl(\hat{R}(\theta) \| R(\theta)) > \varepsilon\} \le (n+1) e^{-n\varepsilon} \quad (9)$$

Equivalently, for any confidence parameter $\delta \in (0,1)$, choosing $\varepsilon = [\ln(n+1) + \ln(1/\delta)]/n$ yields the high-probability statement that with probability at least $1-\delta$:

$$kl(\hat{R}(\theta) \| R(\theta)) \le [\ln(n+1) + \ln(1/\delta)]/n \quad (10)$$

This inequality controls the deviation between empirical and true risk for a fixed parameter $\theta$. The next step lifts this fixed-parameter result to a uniform statement that holds simultaneously for all posterior distributions $Q$. For each $\theta$, define:

$$Z_\theta = n \cdot kl(\hat{R}(\theta) \| R(\theta)) \quad (11)$$

From the moment bound it follows that $\mathbb{E}_S[e^{Z_\theta}] \le n+1$ for every $\theta$, and therefore the prior expectation satisfies $\mathbb{E}_{\theta \sim P} \mathbb{E}_S[e^{Z_\theta}] \le n+1$. The Donsker–Varadhan variational identity states that for any measurable function $f(\theta)$:

$$\ln \mathbb{E}_{\theta \sim P}[e^{f(\theta)}] = \sup_Q \{\mathbb{E}_{\theta \sim Q}[f(\theta)] - KL(Q \| P)\} \quad (12)$$

Applying this identity with $f(\theta) = Z_\theta$ and using $\ln \mathbb{E}_{\theta \sim P}[e^{Z_\theta}] \le \ln(n+1)$ yields:

$$\sup_Q \{\mathbb{E}_{\theta \sim Q}[Z_\theta] - KL(Q \| P)\} \le \ln(n+1) \quad (13)$$

Consequently, for every posterior $Q$ it holds that $\mathbb{E}_{\theta \sim Q}[Z_\theta] - KL(Q \| P) \le \ln(n+1)$, and dividing by $n$ gives the expectation form:

$$\mathbb{E}_{\theta \sim Q}[kl(\hat{R}(\theta) \| R(\theta))] \le [KL(Q \| P) + \ln(n+1)]/n \quad (14)$$

Applying Markov's inequality to the exponentiated left-hand side upgrades this expectation inequality to a high-probability statement:

$$\mathbb{E}_{\theta \sim Q}[kl(\hat{R}(\theta) \| R(\theta))] \le [KL(Q \| P) + \ln((n+1)/\delta)]/n \quad (15)$$

with probability at least $1-\delta$ over the draw of $S$.

Because the binary Kullback–Leibler function $(p, q) \mapsto kl(p \| q)$ is jointly convex, Jensen's inequality yields:

$$kl(\hat{R}(Q) \| R(Q)) \le \mathbb{E}_{\theta \sim Q}[kl(\hat{R}(\theta) \| R(\theta))] \quad (16)$$

Combining the last two displayed inequalities produces the central KL-PAC-Bayes inequality used in this work. Thus, with probability at least $1-\delta$ over $S$, the following holds simultaneously for every posterior $Q$:

$$kl(\hat{R}(Q) \| R(Q)) \le [KL(Q \| P) + \ln((n+1)/\delta)]/n \quad (17)$$

with the binary KL given explicitly by:

$$kl(p \| q) = p\ln(p/q) + (1-p)\ln[(1-p)/(1-q)], \quad p, q \in [0,1] \quad (18)$$

To obtain a computable upper bound on $R(Q)$, define:

$$\varepsilon = KL(Q \| P) + \ln((n+1)/\delta)/n \quad (19)$$

Let $p = \hat{R}(Q)$, and determine the smallest $q \in [p, 1)$ satisfying:

$$p\ln(p/q) + (1-p)\ln[(1-p)/(1-q)] = \varepsilon. \quad (20)$$

The scalar $q$ furnishes an explicit high-probability upper bound $R(Q) \le q$. Because the left-hand side is continuous and strictly increasing in $q$ for $q > p$, the solution $q$ can be found efficiently

by bisection or other root-finding methods. A looser but closed-form alternative is:

$$R(Q) \leq \hat{R}(Q) + \sqrt{[KL(Q \| P) + ln((n+1)/\delta)]/2n} \quad (21)$$

For practical computation in ALL-PET experiments, it is convenient to choose Gaussian prior and posterior distributions. If $Q = \mathcal{N}(\mu_Q, \Sigma_Q)$ and $P = \mathcal{N}(\mu_P, \Sigma_P)$ then:

$$KL(Q \| P) = [ln(det(\Sigma_P)/det(\Sigma_Q)) - d + tr(\Sigma_P^{-1}\Sigma_Q) \\ + (\mu_Q - \mu_P)^\top \Sigma_P^{-1}(\mu_Q - \mu_P)]/2 \quad (22)$$

where $d$ denotes the dimension of the parameter vector $\theta$. In the isotropic special case $\Sigma_P = \sigma_P^2 I$ and $\Sigma_Q = \sigma_Q^2 I$, the expression simplifies to:

$$KL(Q \| P) = [dln(\sigma_P^2)/(\sigma_Q^2) - d + d(\sigma_Q^2)/(\sigma_P^2) \\ + \| \mu_Q - \mu_P \|_2^2 / \sigma_P^2]/2 \quad (23)$$

A convenient and commonly used choice is $\mu_P = 0$, $\mu_Q = \hat{\theta}$ (the flattened checkpoint parameter vector), and $\sigma_Q^2 = \sigma_P^2$, which reduces the KL to the simple form:

$$KL(Q \| P) = \| \theta \|_2^2 / 2\sigma_P^2 \quad (24)$$

The empirical Gibbs risk $\hat{R}(Q)$ is estimated by Monte Carlo sampling from $Q$. Draw $m$ independent parameter samples $\theta_1, \ldots, \theta_m \sim Q$ and compute:

$$\hat{R}(Q) \approx \sum_{j=1}^{m} \hat{R}(\theta_j)/m = \sum_{j=1}^{m} \sum_{i=1}^{n} \ell(\theta_j; y_i)/mn \quad (25)$$

If $Q$ is highly concentrated, the point estimate $\hat{R}(\hat{\theta})$ may be substituted to reduce computation. The Monte Carlo standard error can be controlled by selecting a sufficiently large $m$, typical values in practice are $m \in [100, 500]$ for ALL-PET. Implementation details appropriate for ALL-PET include: selection of the constant $C > 0$ used to rescale an unbounded natural loss into $[0,1]$; explicit choice of prior variance $\sigma_P^2$; specification of posterior mean $\mu_Q$ as the checkpoint $\hat{\theta}$; and numerical inversion of the binary KL to yield the bound $q$. Comparing the numerically obtained upper bounds for models trained before and after augmentation provides a principled, high-confidence assessment of certified generative performance in Fig. 7. For the same confidence levels 0.999, 0.995, 0.99, and 0.95, the bounds after augmentation are much lower. The difference in bounds is about 0.027, which reduces the upper bound of generalization error and improves the model's generalization stability.

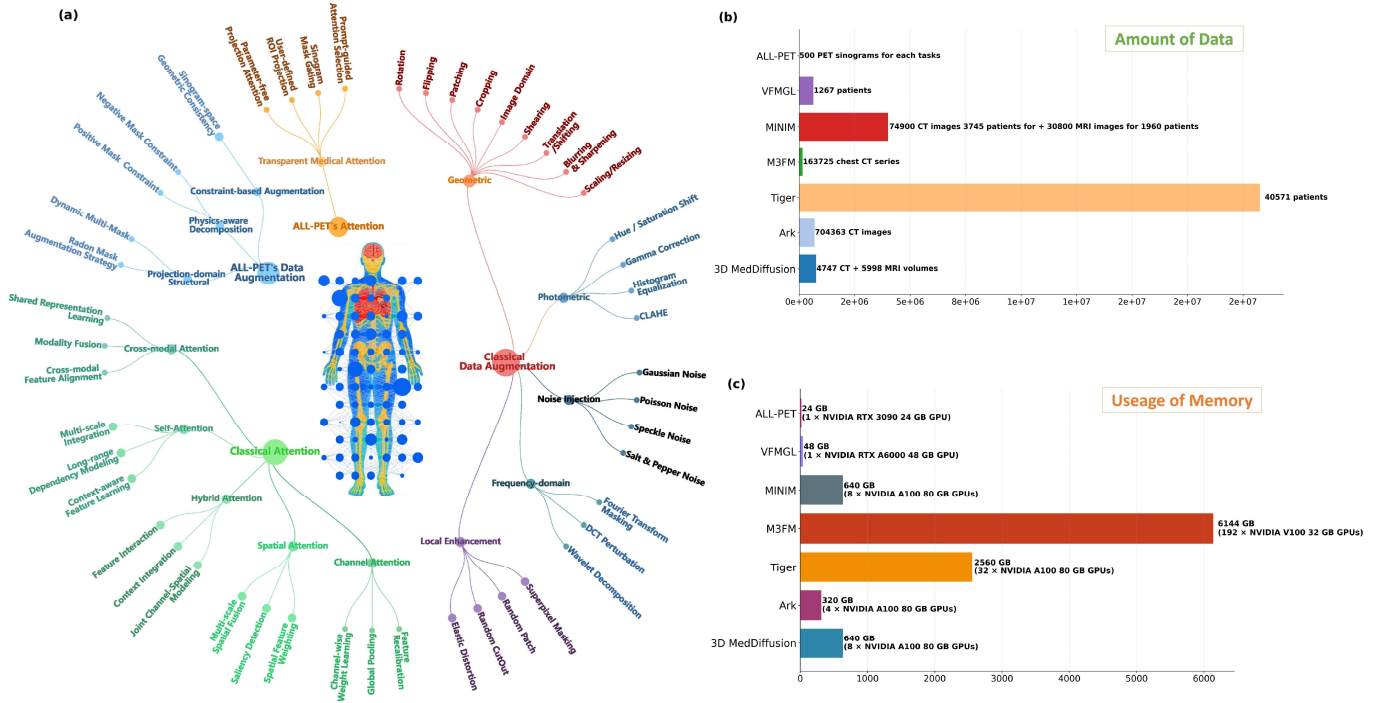

Fig. 8. Architecture design comparison between ALL-PET and classical methods. ALL-PET shifts augmentation and attention to the sinogram space and leverages mask-based physical priors and parameter-free, ROI-driven attention to enable low-shot, low-resource PET modeling. (a) represents different methods between classical ways and ALL-PET's algorithms. (b) represents the amount of data used between existed foundation models and ALL-PET. (c) represents the memory usage between existed foundation models and ALL-PET.

**Discussion**

ALL-PET demonstrates that high-quality generative modeling and downstream PET tasks can be achieved in projection domain with orders of magnitude lower data and compute than typical image domain foundation models. Using as few as 500 sinograms per tracer or anatomical region, ALL-PET attains competitive generation and task performance by combining three complementary design principles: (1) physics-aware data augmentation RMAS that expands training variability without collecting new patient scans; (2) geometry-based, parameter-free regularization DMM and positive/negative decomposition that embeds sinogram structure as an implicit prior; and (3) TMA, a clinician-controllable, non-parametric attention mechanism that highlights diagnostically relevant ROI in sinogram space. Together these elements shift complexity from large parameter counts to intelligent data and constraint design, enabling a low-shot, low-resource foundation model suitable for resource-constrained clinical settings.

Compared to contemporary medical foundation models, such as 3D MedDiffusion[9], Ark[16], Tiger[54], M3FM[8], MINIM[55], and

VFMGL[30] that rely on hundreds of thousands of images and multi-node GPU clusters in Fig. 8, ALL-PET's runtime and memory requirements are substantially reduced: training and inference are feasible on a single RTX 3090 (24 GB) GPU. This reduction is not achieved by sacrificing expressiveness but by increasing the informational content and diversity of the training set through RMAS/DMM and by enforcing mathematically grounded sinogram decompositions. Consequently, ALL-PET maintains or improves PET generation and analysis accuracy while markedly lowering the barrier for deployment in regional hospitals, mobile/portable scanners, and other low-resource environments.

The value of operating in projection domain is twofold. First, sinograms carry explicit geometric information (radial position, angle) that can be systematically manipulated through masks; RMAS exploits this property to produce realistic, structure-preserving augmentations that general image-domain transforms cannot replicate. Second, projection space constraints yield zero-cost regularizers that encode measurement physics directly into training, improving robustness to out-of-distribution scans and reducing the need for large network capacity. TMA further strengthens clinical utility by allowing clinicians to inject domain knowledge (ROI specification or correction) without retraining, making ALL-PET both interpretable and workflow friendly.

We acknowledge several limitations and areas for further validation. First, while 500 samples per task are sufficient in our experiments, the optimal sample size will depend on tracer heterogeneity, scanner hardware, and clinical endpoint; rigorous multi-center validation is required to quantify generalization across vendors and protocols. Second, RMAS/DMM assume that forward-projected image masks produce sinogram variations representative of real anatomical and acquisition variability; edge cases with severe motion, complex scatter, or highly atypical tracer distributions may require task-specific mask priors or additional physics-based simulation. Third, TMA depends on either automated thresholding of low-resolution FBP or clinician-specified ROIs, both approaches introduce user-dependent variability. We recommend deploying calibrated interfaces and logging ROI edits to ensure reproducibility and regulatory traceability in clinical trials.

In summary, ALL-PET offers a practical route to foundation modeling in measurement space by reallocating complexity from large neural parameterizations to physics-informed data design and transparent attention. This paradigm reduces dataset and hardware requirements while preserving interpretability and clinical flexibility, thereby lowering the adoption barrier for advanced generative and analytic PET models in real-world, resource-constrained healthcare environments.

## Acknowledgements
This work received funding from the National Natural Science Foundation of China (62122033, 62201193).

## Author contributions
B.H. and Q.L. conceived the idea. B.H. designed the experiments. B.H. and K.C. performed the data preparation. B.H. analyzed the data. B.H., and K.C. modeled the network and performed the experiments. B.H. and Q.L. supervised the whole study. All authors discussed the results and contributed to the writing.

## Competing financial interests
The authors declare no competing financial interests.

## Methods
### Dynamic Multi-Mask (DMM)
To simulate structural sparsity and improve robustness against incomplete or occluded sinograms, we introduce the DMM strategy. DMM aims to diversify projection domain during training by injecting multiple randomly located spatial occlusions. This exposes the model to variable masking patterns, making it more adaptable to low-shot medical imaging scenarios.

Given a 2D image function $f(x, y)$, its Radon transform $R[f](s, \theta)$ projects the image onto the sinogram space along lines with angle $\theta$ and distance $s$ from the origin:

$$R[f](s,\theta) = \iint f(x, y)\delta(x\cos\theta + y\sin\theta - s)dxdy \qquad (1)$$

In each training iteration, $K$ non-overlapping spatial masks $M_k(x, y)$ are randomly sampled. These masks simulate regional occlusion in the image and are projected to the sinogram domain using Radon transform. Since the Dirac delta function is not directly implementable in the discrete domain, it is approximated by a smooth kernel $K(\cdot)$ interpolation:

$$M_k^S(s,\theta) \approx \sum_{i,j} M_k(x_i, y_j) \cdot K(s - x_i\cos\theta - y_j\sin\theta) \qquad (2)$$

The sinogram domain masks are summed:

$$M^S(s,\theta) = \sum_{k=1}^{K} M_k^S(s,\theta) \qquad (3)$$

To ensure the resulting values remain consistent and bounded, two normalization schemes are applied.

Linear normalization rescales the mask:

$$\hat{M}^S(s,\theta) = \frac{M^S(s,\theta)}{max_{s,\theta} M^S(s,\theta)} \qquad (4)$$

Clipping normalization bounds values into

$$\hat{M}^S(s,\theta) = clip(M^S(s,\theta), 0, 1) \qquad (5)$$

Importantly, the Radon transform is linear. For disjoint image masks, we maintain:

$$R[\sum_k f_k(x, y)] = \sum_k R[f_k(x, y)] \qquad (6)$$

This property preserves structural decomposability and interpretability in sinogram - space augmentations.

### Radon Mask Augmentation (RMA)
RMA introduces supervision in both masked and unmasked sinogram regions to encourage localized feature learning. Starting from the normalized sinogram mask $M^S(s,\theta)$, we construct two complementary sinogram data:

Positive data captures regions preserved by the mask:

$$C_{pos}(s,\theta) = M^S(s,\theta) \cdot S(s,\theta) \qquad (7)$$

Negative data represents the removed regions:

$$C_{neg}(s,\theta) = (1 - M^S(s,\theta)) \cdot S(s,\theta) \qquad (8)$$

where $S(s,\theta)$ is the original full sinogram. These data enable the network to learn different representations depending on signal presence. To guide this decomposition, we define the positive-negative consistency loss:

$$\mathcal{L}_{PN} = \| C_{pos} - M^S \cdot S \|^2 + \| C_{neg} - (1 - M^S) \cdot S \|^2 \qquad (9)$$

This regularization encourages feature consistency with respect to masked structure and suppresses spurious activation in irrelevant regions.

### Transparent Medical Attention (TMA)

TMA leverages pre-segmented lesion regions to guide sinogram attention without introducing additional trainable parameters. This enhances the interpretability of attention mechanisms and strengthens localization.

**Step 1: ROI Extraction**

A coarse image $X(x, y)$ is reconstructed using FBP or MLEM method. A binary lesion mask $M_{ROI}^I(x, y)$ is extracted by applying a global or adaptive threshold $T$:

$$M_{ROI}^I(x, y) = \begin{cases} 1, & X(x,y) > T \\ 0, & \text{otherwise} \end{cases} \quad (10)$$

**Step 2: ROI Projection to Sinogram**

The binary mask is transformed to projection space using Radon transform:

$$M_{ROI}^S(s, \theta) = R[M_{ROI}^I(x, y)] \quad (11)$$

This represents the accumulation of lesion responses along rays intersecting $M_{ROI}^I$.

**Step 3: Mask-Gated Attention Weighting**

The attention mechanism applies direct element-wise weighting of sinogram features using the projected lesion mask:

$$O(s, \theta) = M_{ROI}^S(s, \theta) \cdot F(s, \theta) \quad (12)$$

where $F(s, \theta)$ is the feature map at a given layer. This suppresses background responses and highlights lesion-related projections in a lightweight and interpretable way.

**Unified Diffusion Integration**

We embed the masked sinogram $S \in \mathbb{R}^{3 \times H \times W}$ into a latent variable $z_T$ using a sinogram autoencoder. The latent diffusion model[1] is adapted to different diffusion samplers to accommodate various upstream and downstream tasks. The reconstruction process employs progressive denoising guided by DDPM[2], DDIM[3] and DDBM[4] strategies

The sinogram data augmented by RMA strategy is concatenated to form the composite input:

$$C_{all} = [C_{pos}, S_{full}, C_{neg}] \in \mathbb{R}^{3 \times H \times W} \quad (13)$$

where $H$ and $W$ correspond to the sinogram height and width, respectively. An encoder $E$ projects $C_{all}$ into a latent code:

$$z_0 = E(C_{all}) \in \mathbb{R}^d \quad (14)$$

which serves as the starting point for all subsequent discrete diffusion processes.

**For DDPM sampler:**

A forward corruption chain is defined for $t = 1, \ldots, T$ by:

$$q(z_t | z_{t-1}) = \mathcal{N}(z_t; \sqrt{\alpha_t} z_{t-1}, (1-\alpha_t)\mathbf{I}), \quad \alpha_t = 1 - \beta_t, \quad (15)$$

with the cumulative product $\bar{\alpha}_t = \prod_{i=1}^t \alpha_i$. Equivalently, one obtains the closed-form marginal:

$$q(z_t | z_0) = \mathcal{N}(z_t; \sqrt{\bar{\alpha}_t} z_{t-1}, (1-\bar{\alpha}_t)\mathbf{I}). \quad (16)$$

The learned reverse transform is parameterized by a neural network $\epsilon_\theta$ that predicts the noise component. Denoting $\sigma_t^2 = \frac{1 - \bar{\alpha}_{t-1}}{1 - \bar{\alpha}_t} \beta_t$, the reverse distribution is:

$$p_\theta(z_{t-1} | z_t) = \mathcal{N}(z_{t-1}; \mu_\theta(z_t, t), \sigma_t^2 \mathbf{I}), \quad (17)$$

with mean:

$$\mu_\theta(z_t, t) = [z_t - (1 - \alpha_t)\epsilon_\theta(z_t, t) / \sqrt{1-\bar{\alpha}_t}] / \sqrt{\alpha_t} \quad (18)$$

The denoising objective minimizes the expected squared error in noise prediction:

$$\mathcal{L}_{denoise} = \mathbb{E}_{z_0, \epsilon, t} \| \epsilon - \epsilon_\theta(\sqrt{\bar{\alpha}_t} z_0 + \sqrt{1-\bar{\alpha}_t} \epsilon, t) \|^2. \quad (19)$$

**For DDIM sampler:**

To accelerate inference, a non-Markovian sampling rule is employed. For each timestep $t$, let $\hat{\epsilon} = \epsilon_\theta(z_t, t)$, $\bar{\alpha}_t = \prod_{i=1}^t \alpha_i$, $\bar{\alpha}_{t-1} = \prod_{i=1}^{t-1} \alpha_i$.

Then the update to $z_{t-1}$ reads

$$z_{t-1} = \sqrt{\bar{\alpha}_{t-1}}(z_t - \sqrt{1-\bar{\alpha}_t}\hat{\epsilon})/\sqrt{\bar{\alpha}_t} + \sqrt{1 - \bar{\alpha}_{t-1} - \eta^2}\hat{\epsilon} + \eta \mathbf{u}, \quad (20)$$
$$\mathbf{u} \sim \mathcal{N}(0, \mathbf{I}),$$

where $\eta \in [0, 1]$ controls stochasticity. When $\eta = 0$, sampling is entirely deterministic:

$$z_{t-1} = \sqrt{\bar{\alpha}_{t-1}}(z_t - \sqrt{1-\bar{\alpha}_t}\hat{\epsilon})/\sqrt{\bar{\alpha}_t} + \sqrt{1 - \bar{\alpha}_{t-1}}\hat{\epsilon}. \quad (21)$$

This scheme preserves the same marginals $q(z_t | z_0)$ while permitting significantly fewer steps.

**For DDBM sampler:**

In contrast to conventional samplers, we adopt a denoising diffusion bridge mechanism to generate the latent variable $z_0$ from the terminal state $z_t$ under a bridge prior.

Specifically, the forward stochastic differential equation (SDE) is defined as:

$$dz_t = f(z_t, t)dt + g(t)^2 \nabla_{z_t} \log p(z_T | z_t) dt + g(t) dw_t \quad (22)$$

where $f(z_t, t)$ is the forward drift term (e.g., 0 in VE-type diffusion, or $\frac{d \log \alpha_t}{dt} z_t$ in VP-type), $g(t)$ is the noise scale, and $\nabla_{z_t} \log p(z_T | z_t)$ is the Doob's h-transform representing a bridge constraint toward $z_t$.

To generate samples, the reverse-time SDE is constructed as:

$$dz_t = [f(z_t, t) - g(t)^2(s_\theta(z_t, z_T, t) - \nabla_{z_t} \log p(z_T | z_t))]dt + g(t)d\hat{w}t \quad (23)$$

where $s_\theta$ is the denoising bridge score predicted by a time-aware latent diffusion model conditioned on both current and terminal latent states.

When using deterministic inference, the bridge-based probability flow ODE is employed:

$$dz_t / dt = f(z_t, t) - g(t)^2(s_\theta(z_t, z_T, t)/2 - \nabla_{z_t} \log p(z_T | z_t)) \quad (24)$$

To approximate the bridge score $s_\theta$, we adopt a reparameterization scheme based on predicted $z_{0|t}$ given by:

$$s_\theta(z_t, z_T, t) = -\frac{z_t - (\frac{SNR_T}{SNR_t} \frac{\alpha_t}{\alpha_T} z_T + \alpha_t \cdot D_\theta(z_t, z_T, t) \cdot (1 - \frac{SNR_T}{SNR_t}))}{\sigma_t^2(1 - \frac{SNR_T}{SNR_t})} \quad (25)$$

where $\alpha_t, \sigma_t$ are the signal and noise coefficients, $SNR_t = \alpha_t^2 / \sigma_t^2$, and $D_\theta$ is the predicted $z_0$ from the latent denoising model. This score expression allows bridge-guided supervision even during inference.

This integration of denoising diffusion, implicit sampling, and bridge supervision upon the physically constrained augmented sinogram input yields a versatile foundation model for both unconditional sinogram synthesis and task-adapted PET downstream processing.

## Data availability

The synthetic data of ALL-PET can be obtained from: https://github.com/yqx7150/ALL-PET.

## Code availability

The code used in this study is publicly available at: https://github.com/yqx7150/ALL-PET.

**Appendix:**

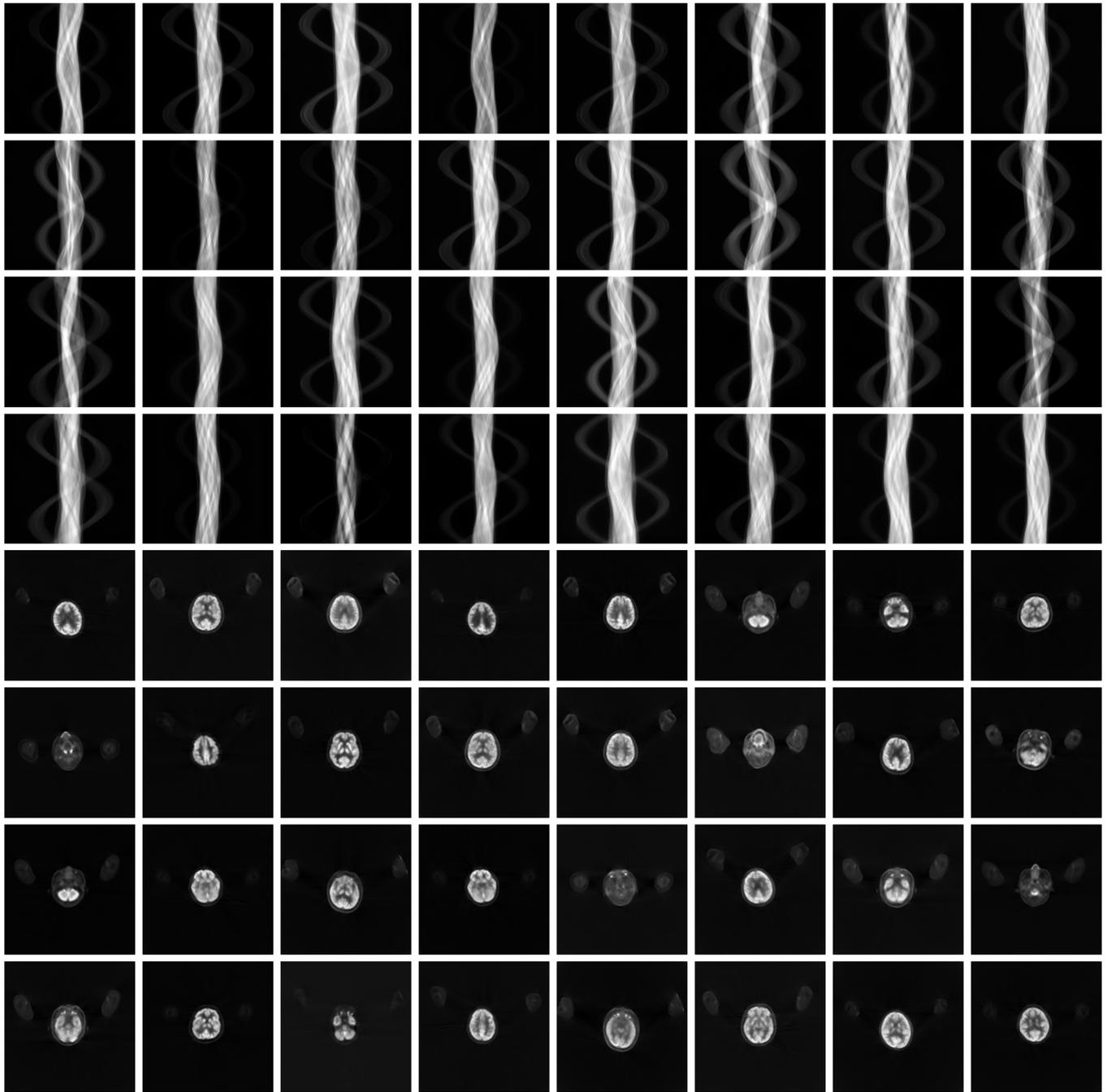

Fig. 9. Synthetic sinograms and their corresponding reconstructed images for "$^{18}$F-FDG" & "Brain" from ALL-PET.

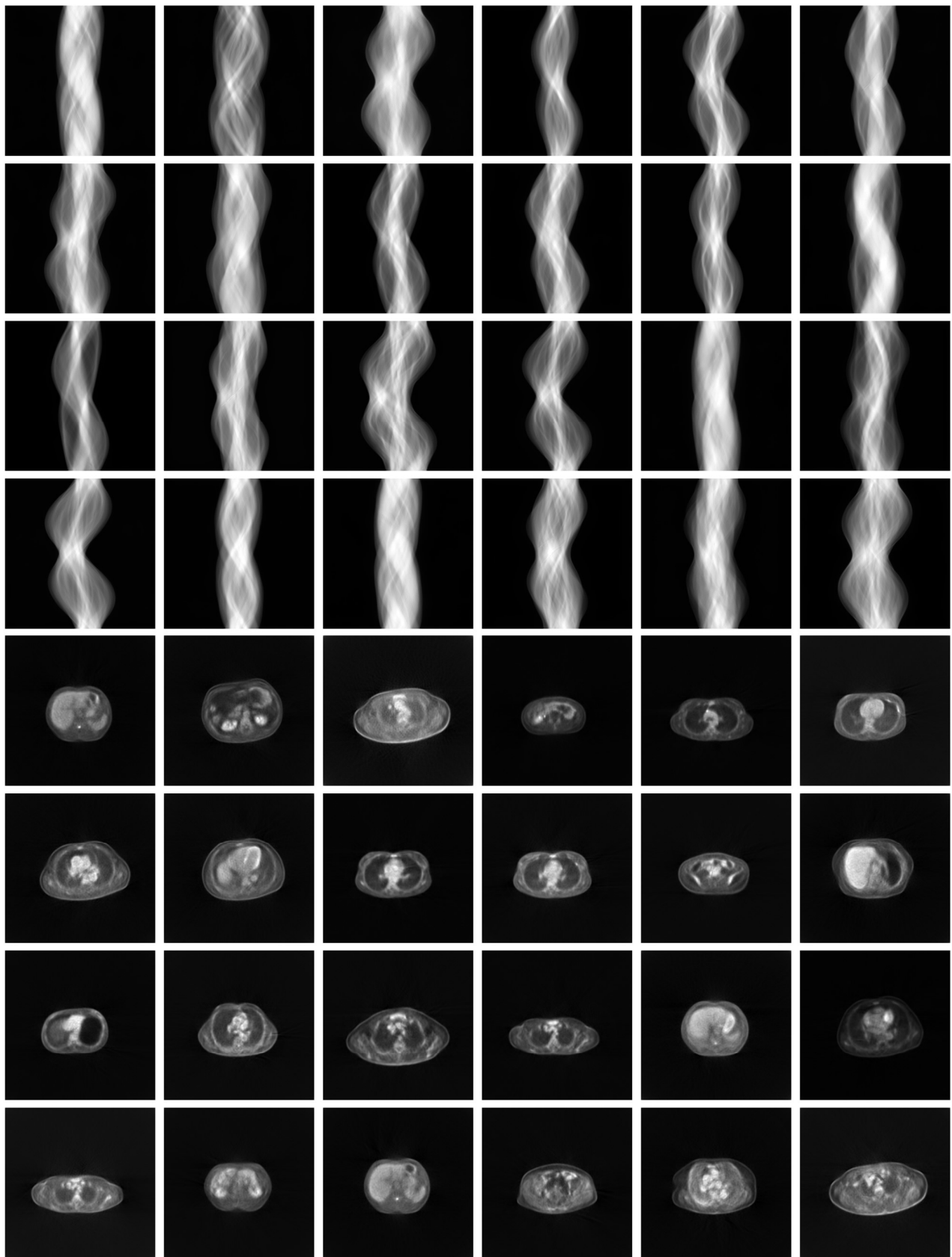

Fig. 10. Synthetic sinograms and their corresponding reconstructed images for "$^{18}$F-FDG" "Trunk" from ALL-PET.

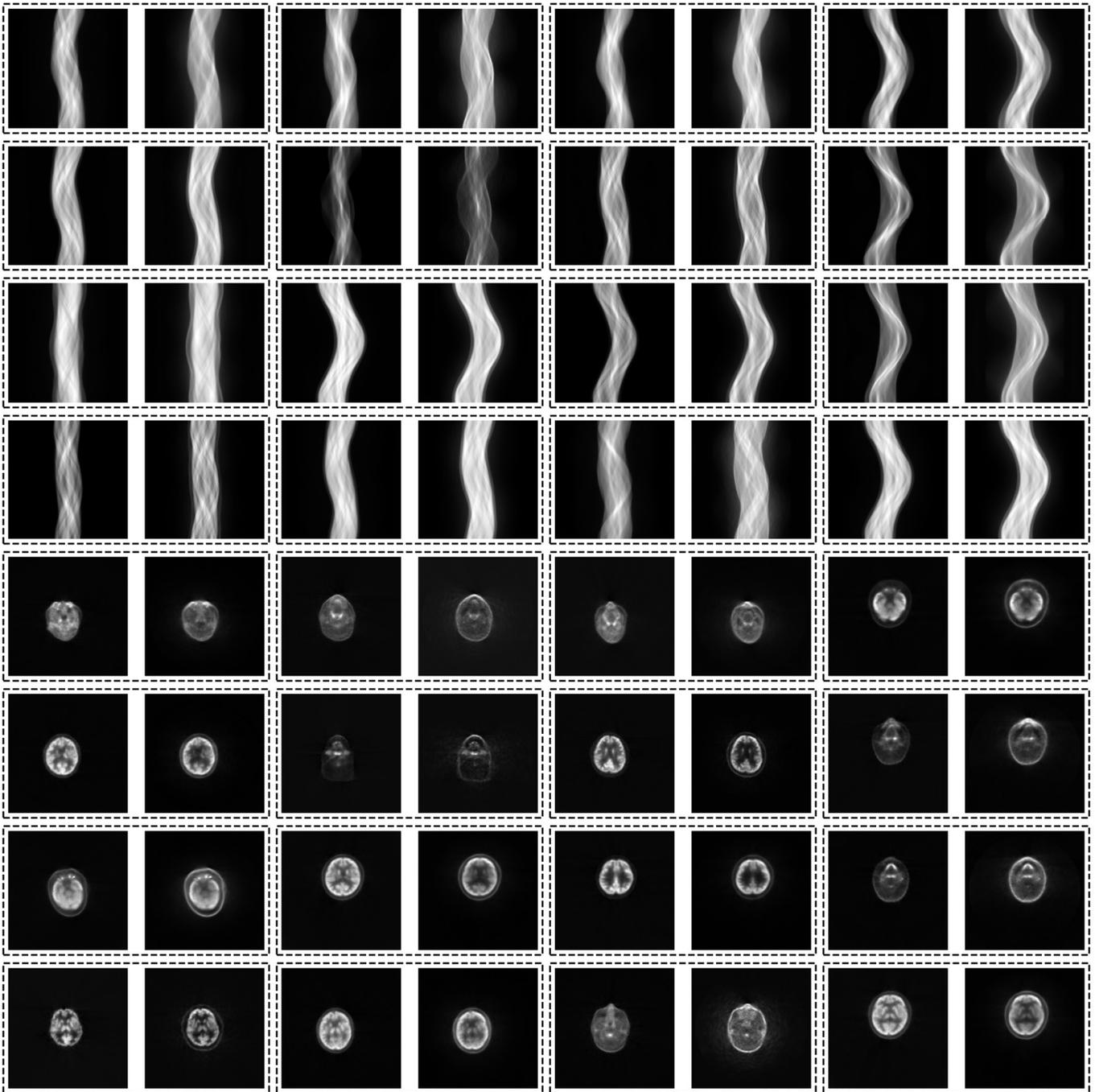

Fig. 11. Paired synthetic sinograms and their corresponding reconstructed images for "Attenuation artifact correction" from ALL-PET. Each pair of data is enclosed in a dashed box; within each box, the left panel represents the data after self-attenuation correction, and the right panel represents the data before self-attenuation correction.

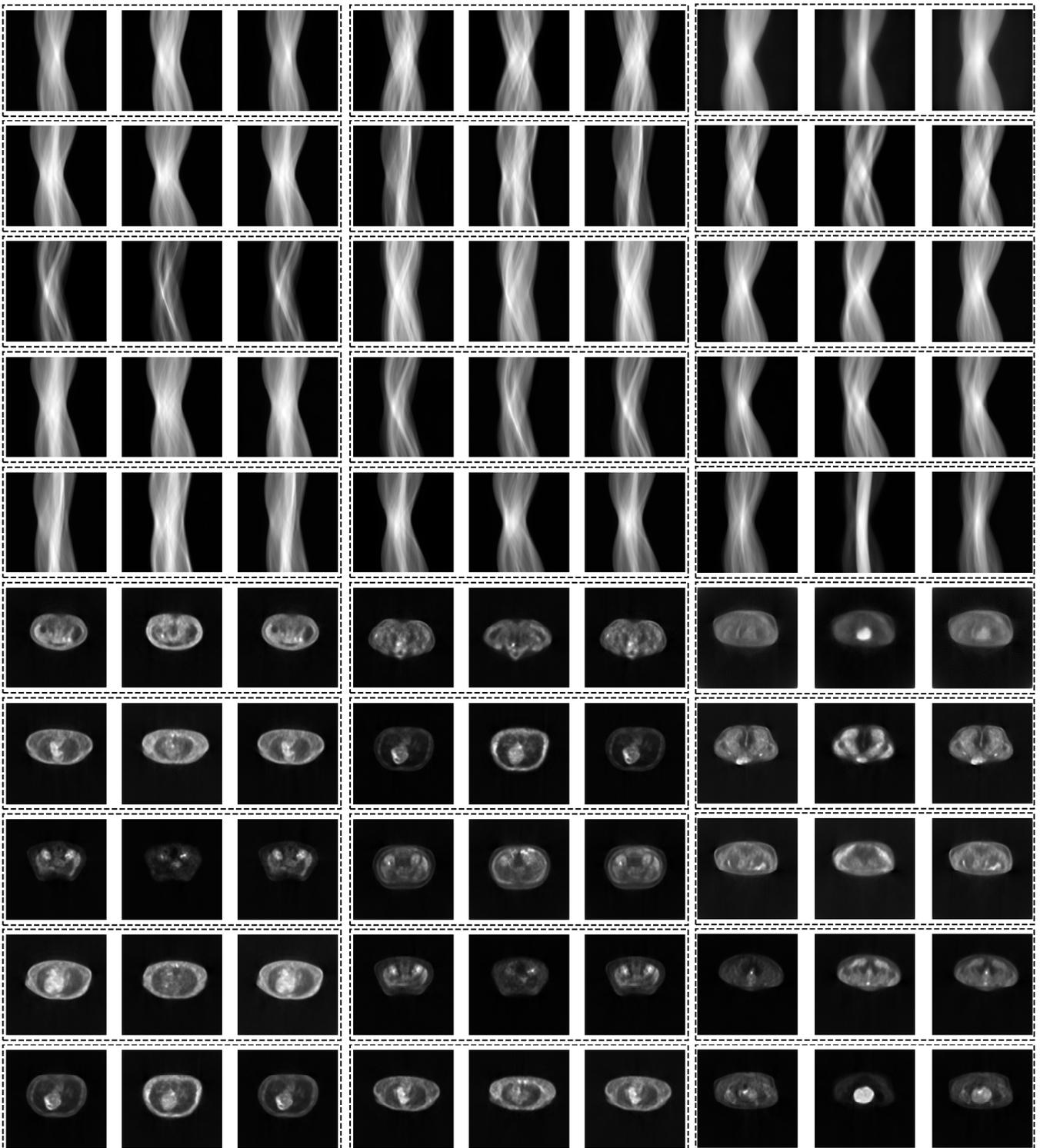

Fig. 12. Paired synthetic sinograms and their corresponding reconstructed images for "Tracer separation" from ALL-PET. Each pair of data is enclosed in a dashed box; within each box, the left panel represents the dual-tracer data, the middle and right panels represent the separated tracers.

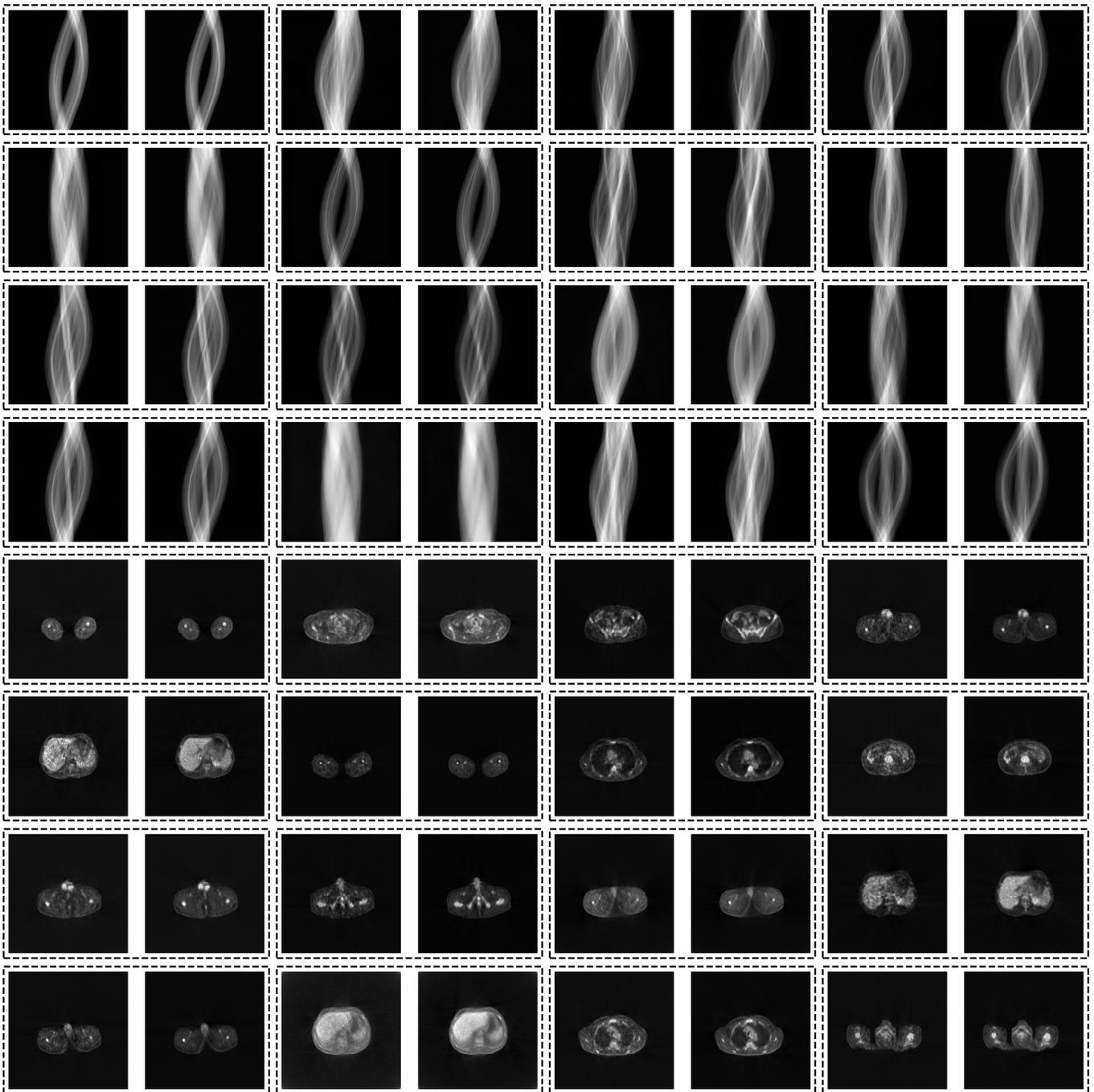

Fig. 13. Paired synthetic sinograms and their corresponding reconstructed images for "Low-dose reconstruction" from ALL-PET. Each pair of data is enclosed in a dashed box; within each box, the left panel represents the 1/4 low-dose data, and the right panel represents the normal-dose data.

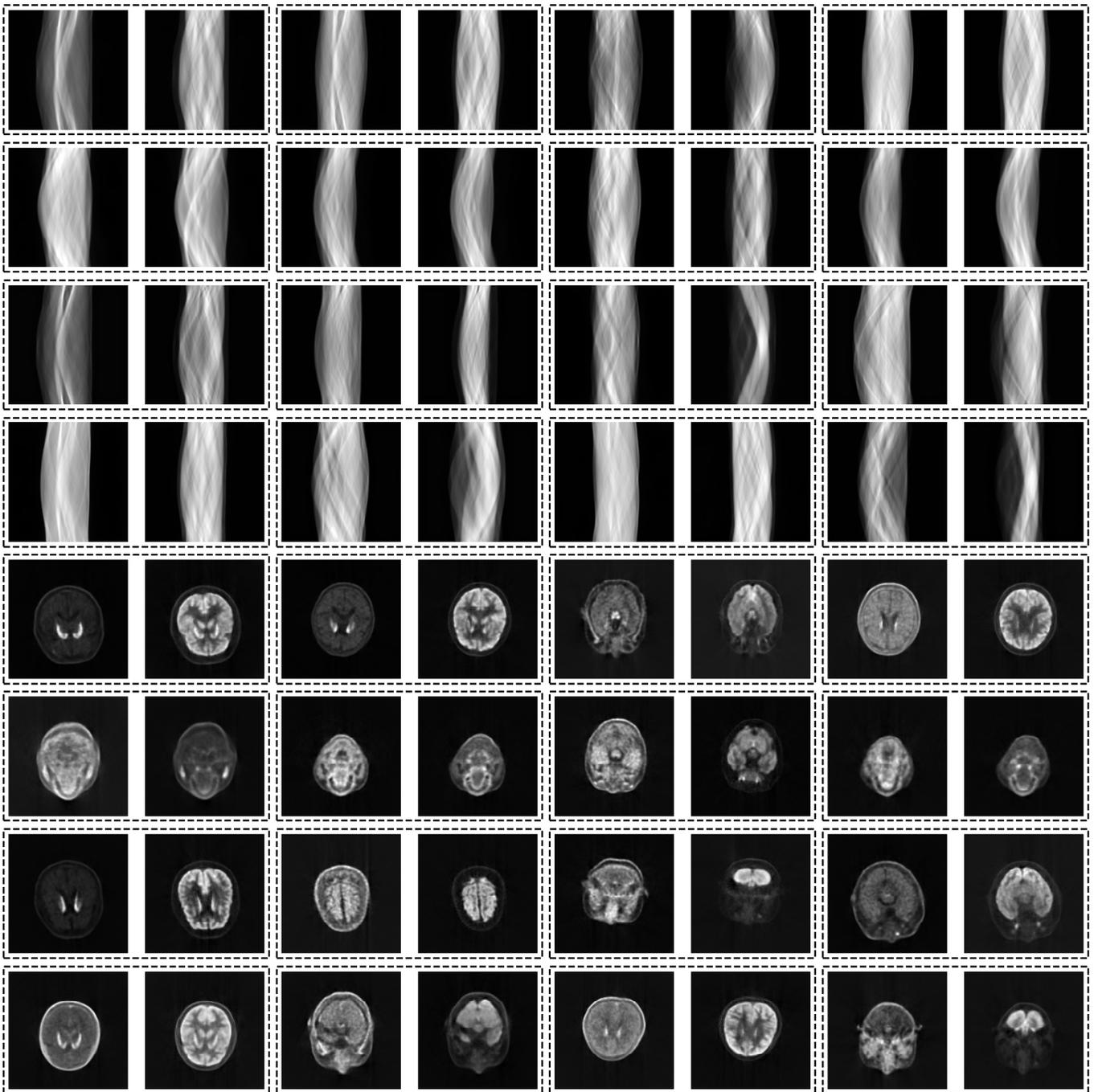

Fig. 14. Paired synthetic sinograms and their corresponding reconstructed images for "Tracer conversion" from ALL-PET. Each pair of data is enclosed in a dashed box; within each box, the left and right panel represents the different tracers.